# Self-Supervised Online Camera Calibration for Automated Driving and Parking Applications

Ciarán Hogan[1,2], Ganesh Sistu[1], Ciarán Eising[2]

[1]*Valeo Vision Systems Galway, Ireland*

[2]*Department of Electronic and Computer Engineering University of Limerick Ireland*

**Abstract**

Camera-based perception systems play a central role in modern autonomous vehicles. These camera-based perception algorithms require an accurate calibration to map the realworld distances to image pixels. In practice, calibration is a laborious procedure requiring specialised data collection and careful tuning. This process must be repeated whenever the parameters of the camera change, which can be a frequent occurrence in autonomous vehicles. Hence there is a need to calibrate at regular intervals to ensure the camera is accurate. Proposed is a deep learning framework to learn intrinsic and extrinsic calibration of the camera in real time. The framework is self-supervised and doesn't require any labelling or supervision to learn the calibration parameters. The framework learns calibration without the need for any physical targets or to drive the car on special planar surfaces.

## 1 Introduction

Typically, cameras for autonomous vehicles must be calibrated intrinsically and extrinsically. Intrinsic parameters relate to the camera's internal factors like focal length and optical centre. Extrinsic parameters relate to external factors like location and orientation. The intrinsic calibration is typically done during the camera manufacturing stage and is done in specialised lab settings. The extrinsic calibration is typically done at regular intervals to make sure the camera is performing adequately. The Intrinsic values of the camera also change over time due to factors such as temperature, humidity, general wear and tear and mechanical alignment changes due to continuous vibrations. Hence the proposed deep learning method cuts out the need for specialized lab conditions or patterns and can learn calibration in real time over the lifetime of the vehicle.

Traditional methods of camera calibration usually require specialised lab settings and some kind of target pattern like chessboards or April tags. Introduced in 2000 Zhangs method [Zhang, 2000] has become one of the most popular camera calibration methods in the world. The method involves detecting feature points of a pattern under different orientations by moving either the camera or plane.

While camera calibration has traditionally relied on geometric and statistical methods mentioned, deep learning has opened up new possibilities for calibrating cameras. [Bogdan et al., 2018] uses a deep learning approach for automatic intrinsic calibration of wide field-of-view cameras.

Properties of Self supervised depth and pose estimation networks like [Guizilini et al., 2020] and [Godard et al., 2019] can be used a proxy to also learn calibration. [Garg et al., 2016] first introduced the idea of the joint learning of depth and ego-motion. The proposed method provided a significant contribution to the subject of depth estimation as it allows for the learning of depth estimation from monocular images totally unsupervised. [Fang et al., 2022] uses self supervised depth and ego-motion as a proxy to learn calibration in parallel. The architecture consists of a self-supervised depth and ego-motion framework that provides End-to-End Self-Calibration.

# 2 Methodology

The proposed calibration framework uses self-supervised monocular depth and pose estimation as a proxy for learning camera calibration with the addition of a third network to learn intrinsics. Self-supervised depth and ego-motion architectures consist of a depth network that produces a depth map and a pose network that predicts the transformation between the current frame and the context frame. With this known transformation and depth map, one can warp/project the current frame into a target image and train the networks jointly by minimising the photometric loss between the actual image and the synthesised image from the projection. Monodepth2 [Godard et al., 2019] was chosen as a base framework for the project. Monodepth2 is a popular and well documented Pytorch-based self-supervised monocular depth estimation network. At the core of the project is a depth network with a multi-input ResNet [He et al., 2016] encoder and UNet [Ronneberger et al., 2015] decoder alongside a pose CNN, also with a multi-input ResNet encoder.

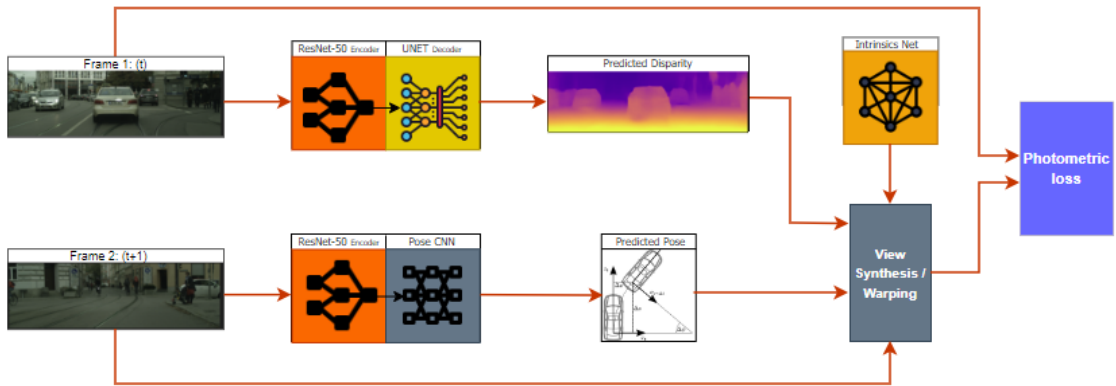

Figure 1: Calibration Framework Architecture

## 2.1 Architecture

The modified monodepth2 framework consists of:
- Depth network: ResNet-50 encoder pre-weighted on ImageNet & UNet decoder.
- Pose network: Multi-input ResNet-50 encoder pre-weighted on ImageNet & Pose CNN.
- Intrinsic network consists of 4 or more trainable parameters (Depending on the camera model) representing intrinsic camera parameters which feed into the view synthesis function for image warping.

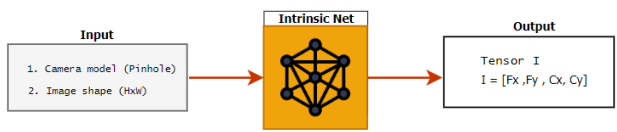

$$\mathbf{K} = \begin{pmatrix} f_x & 0 & c_x \\ 0 & f_y & c_y \\ 0 & 0 & 1 \end{pmatrix}$$

(a) Intrinsic Network pseudo diagram      (b) Traditional Intrinsic matrix

### 2.1.1 Intrinsic Network

The intrinsic network consists of 4 or more trainable parameters (Depending on camera model) which represent the intrinsic parameters of the relevant camera model. The network takes two inputs, the camera model and the image shape. The network outputs a tensor of size 1x4 (depending on camera model). This tensor is then manipulated into a traditional 3x3 intrinsic matrix which is fed into the image warping function and the network is trained in simultaneously with the depth and pose network by minimising the photometric loss.

### 2.1.2 Data

The framework was trained on the KITTI dataset. The KITTI dataset is a benchmark dataset commonly used in computer vision for depth estimation. The Eigen zhou split was used which is a subset of the KITTI dataset which consists of 39K images for training and 4k images for validation. A further 679 unseen images are used for evaluation.

## 3 Training/Experiment and Results

The framework is trained in a self-supervised manner by minimising photometric reconstruction error. The framework was trained for 20 Epochs with a learning rate of 0.0001 and a batch size of 4. Training took 36 hours on an NVIDIA GeForce RTX 2080ti graphics card.

## 4 Results

| Model | Abs Rel↓ | Sq Rel↓ | RMSE↓ | RMSE log↓ |
|---|---|---|---|---|
| Base monodepth2 | 0.114 | 0.931 | 4.810 | 0.192 |
| monodepth2 x IntrinsicNet | 0.112 | 0.851 | 4.706 | 0.188 |

Table 1: Depth metric results vs baseline

Comparing the depth metrics of the baseline vs learned intrinsics is one way to evaluate the learned intrinsics in the absence of synthetic data as we are using depth estimation as a proxy to learn camera calibration. Comparing the two we see that the modified framework has a slight improvement in depth evaluation metrics, possibly due to the learned intrinsics being more accurate than the static KITTI parameters.

| Model | Intrinsics | Abs Rel↓ | Sq Rel↓ | RMSE↓ |
|---|---|---|---|---|
| Gordon et al | K (P) | 0.129 | 0.982 | 5.23 |
|  | L (P) | 0.128 | 0.959 | 5.23 |
| J. Fang et al | L (P) | 0.129 | 0.8393 | 4.96 |
| Monodepth2 | K (P) | 0.114 | 0.931 | 4.810 |
| Monodepth2x IntrinsicNet | L (P) | 0.112 | 0.851 | 4.70 |

Table 2: Depth metric results vs baseline

Table 4 shows a comparison of depth metrics from other camera based learning methods [Fang et al., 2022, Gordon et al., 2019] trained on the Eigen split of the KITTI dataset. K denotes known intrinsics and L denotes learned intrinsics. Note the proposed implementation is trained on the Eigen Zhou split which contains the same files but has extra data added from more challenging scenes and lower lighting. This means the Eigen Zhou split is more challenging and provides a better estimation of the model's generalisation capability. The table provides a quick comparison of improved metrics between baseline and learned intrinsics of other implementations.

# 5 Conclusion & Future work

Proposed was a self-supervised deep learning framework that learns camera calibration from video input. As seen the framework learns calibration quite well and can help provide better depth evaluation metrics by providing more accurate calibration parameters than static calibration parameters provided in KITTI. The proposed framework could potentially be very effective as a recalibration tool for vehicle perception systems and could cut the cost of recalibration for both manufacturer and consumer.

Future work would be to train the framework on synthetic data with known true intrinsics and adapt the framework to work with other camera models e.g the fisheye camera model.